\documentclass[10pt,twocolumn,letterpaper]{article}

\usepackage{multirow}
\usepackage{comment}
\usepackage{tabularx}
\usepackage{array}
\usepackage{siunitx}
\usepackage{wrapfig}

\usepackage{amssymb}
\usepackage{booktabs}

\usepackage{graphicx}
\usepackage{amsmath}
\usepackage{amssymb}
\usepackage{booktabs}
\usepackage{tabularx}
\usepackage{tabu}
\usepackage{soul}

\usepackage{cvpr}              

\definecolor{cvprblue}{rgb}{0.21,0.49,0.74}
\usepackage[pagebackref,breaklinks,colorlinks,allcolors=cvprblue]{hyperref}

\def\paperID{17} 
\def\confName{EarthVision}
\def\confYear{2025}

\title{Bridging Classical and Modern Computer Vision:\\ PerceptiveNet for Tree Crown Semantic Segmentation}

\author{Georgios Voulgaris\\
University of Oxford\\
{\tt\small georgios.voulgaris@biology.ox.ac.uk}
}

\begin{document}
\maketitle
\begin{abstract}
The accurate semantic segmentation of tree crowns within remotely sensed data is crucial for scientific endeavours such as forest management, biodiversity studies, and carbon sequestration quantification. However, precise segmentation remains challenging due to complexities in the forest canopy, including shadows, intricate backgrounds, scale variations, and subtle spectral differences among tree species. Compared to the traditional methods, Deep Learning models improve accuracy by extracting informative and discriminative features, but often fall short in capturing the aforementioned complexities.

To address these challenges, we propose PerceptiveNet, a novel model incorporating a Logarithmic Gabor-parameterised convolutional layer with trainable filter parameters, alongside a backbone that extracts salient features while capturing extensive context and spatial information through a wider receptive field. We investigate the impact of Log-Gabor, Gabor, and standard convolutional layers on semantic segmentation performance through extensive experimentation. Additionally, we conduct an ablation study to assess the contributions of individual layers and their combinations to overall model performance, and we evaluate PerceptiveNet as a backbone within a novel hybrid CNN-Transformer model. Our results outperform state-of-the-art models, demonstrating significant performance improvements on a tree crown dataset while generalising across domains, including two benchmark aerial scene semantic segmentation datasets with varying complexities.
\end{abstract}    
\section{Introduction}
\label{sec:intro}

\begin{figure}[t!]
    \begin{tabularx}{\columnwidth}{
        *{4}{>{\centering\arraybackslash}X} 
    }
        \textbf{Original} &
        \textbf{Mask} &
        \textbf{ResUNet} &
        \textbf{PerceptiveNet}\\
        \begin{subfigure}{\linewidth}
            \centering
            \includegraphics[width=1.2\linewidth]{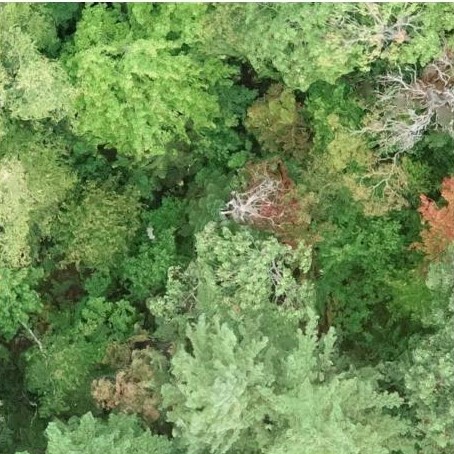}
        \end{subfigure} &
        \begin{subfigure}{\linewidth}
            \centering
            \includegraphics[width=1.2\linewidth]{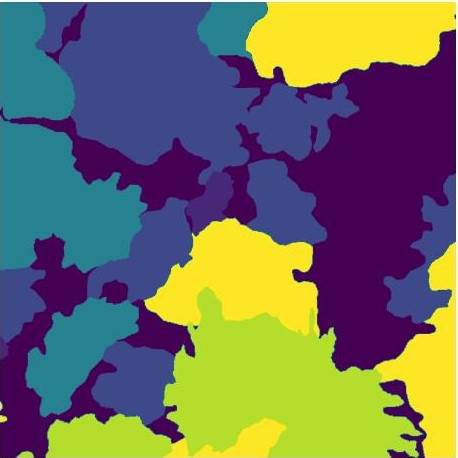}
        \end{subfigure} &
        \begin{subfigure}{\linewidth}
            \centering
            \includegraphics[width=1.2\linewidth]{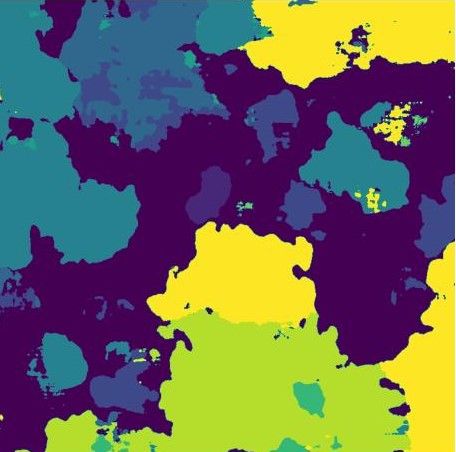}
        \end{subfigure} &
        \begin{subfigure}{\linewidth}
            \centering
            \includegraphics[width=1.2\linewidth]{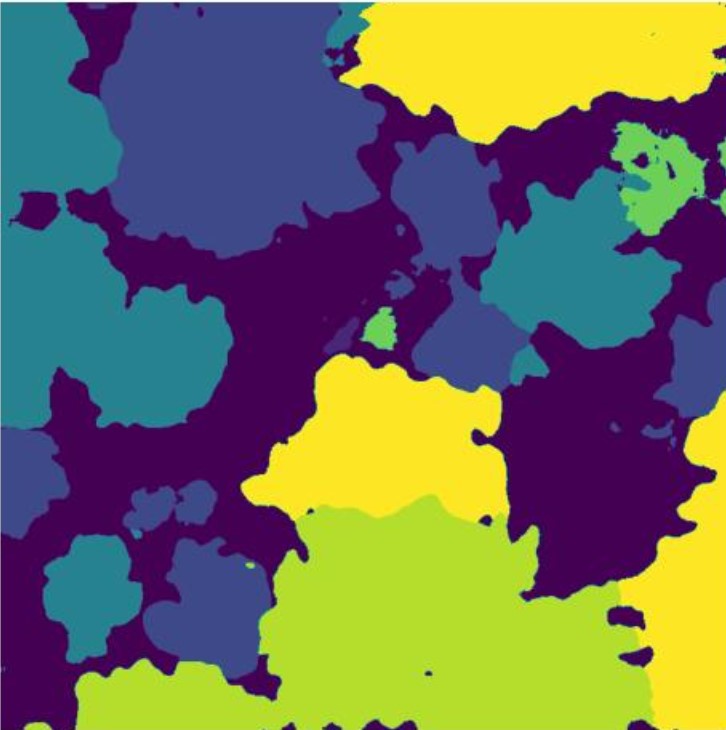}
        \end{subfigure}
    \end{tabularx}
    \vspace{-10pt}
    \caption{Tree Crown Semantic Segmentation, depicting (from left to right): Original Image; Corresponding Mask; ResUNet; and proposed model's semantic segmentation. Each colour represents a different tree species. The images portray densely packed trees with complex boundaries due to partial overlap. Moreover, tree crown similarities, complex occlusion, combined with light variations and shadows further augment the scene's complexity.}
    \label{fig:SemSegexample}
    \vspace{-15pt}
\end{figure}

Tree crowns are crucial indicators of tree health, directly impacting photosynthesis, transpiration, and nutrient absorption (Bakalo et al. \cite{bokalo2013validation}; Li et al. \cite{li2020development}). Accurate semantic segmentation of tree crowns (Figure \ref{fig:SemSegexample}) enables quantitative analysis of forest characteristics, supporting applications in biodiversity assessment, carbon sequestration measurement, and ecological monitoring.

\begin{figure}
    \centering
    \includegraphics[width=0.7\columnwidth]{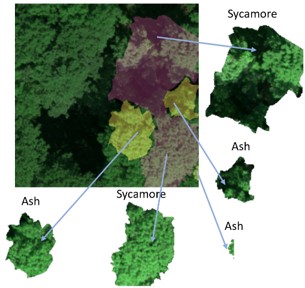}
    \vspace{-8pt}
    \caption{Dense forest canopy, demonstrating the impact of shadows, light  variations, overlapping tree crowns, and weak distinctive features among tree species on the tree crown segmentation.}
    \vspace{-16.3pt}
    \label{fig:SemIssue}
\end{figure}

Dense forests present unique challenges for semantic segmentation: (1) variable lighting conditions and shadows that obscure crown boundaries (Bargas et al. \cite{g2020tree}); (2) complex backgrounds with overlapping canopies (Hu et al. \cite{hu2022uav}); (3) scale variations from different Unmanned Aerial Vehicle (UAV) heights (Ocer et al. \cite{ocer2020tree}); and (4) subtle spectral differences between species (Cao and Zhang \cite{cao2020improved}). Figure \ref{fig:SemIssue} demonstrates these challenges, showing how lighting conditions affect crown appearance of different species.

Current aerial imagery segmentation largely relies on U-Net variants (Ronneberger et al. \cite{ronneberger2015u}), ranging from basic implementations (Ye et al. \cite{ye2022extraction}; Zhang et al. \cite{zhang2020segmenting}; Cao and Zhang \cite{cao2020improved}; Wanger et al. \cite{wagner2019using}; Schiefer et al. \cite{schiefer2020mapping}) to transformer-enhanced architectures (Scheibenreif et al. \cite{scheibenreif2022self}; Vinod et al. \cite{vinod2024novel}; Alshammari and Shahin \cite{alshammari2022efficient}). While specialised modules like dilated convolutions and pyramid pooling (He et al. \cite{he2015spatial}; Zhao et al. \cite{zhao2017pyramid}) improve performance, these approaches fail to address the unique spatial and spectral characteristics of aerial forest scenes. One thing that all the aforementioned works have in common is that they do not consider the challenges that tree crown semantic segmentation in dense forests presents. These challenges translate into specific issues for computer vision algorithms and principles. Additionally, they often ignore the Deep Learning models’ characteristics, feature extraction capabilities, and inherent biases.

\begin{figure*}[!htbp]
    \centering
    \includegraphics[width=0.8\textwidth]{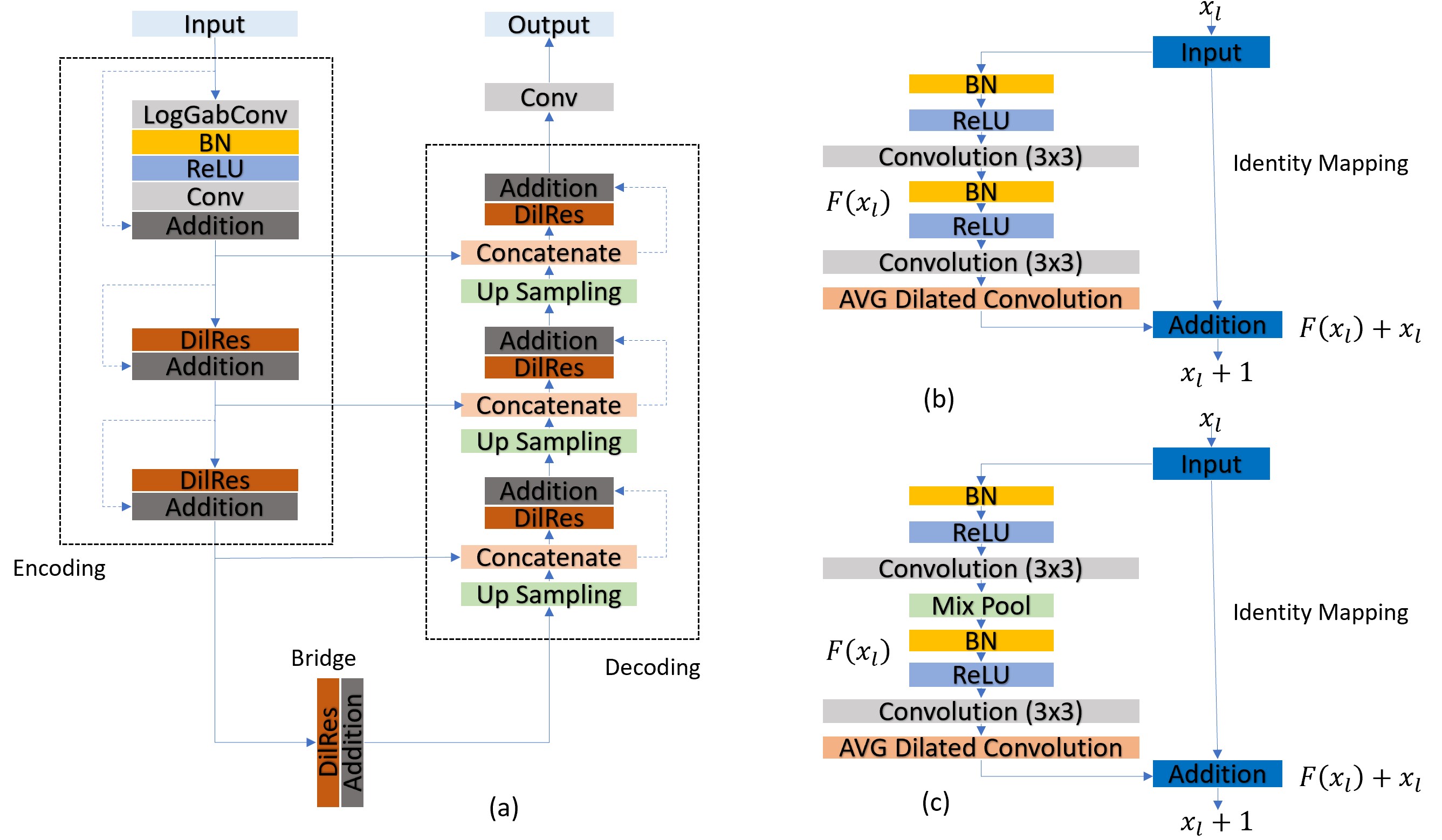}
    \includegraphics[width=0.19\textwidth]{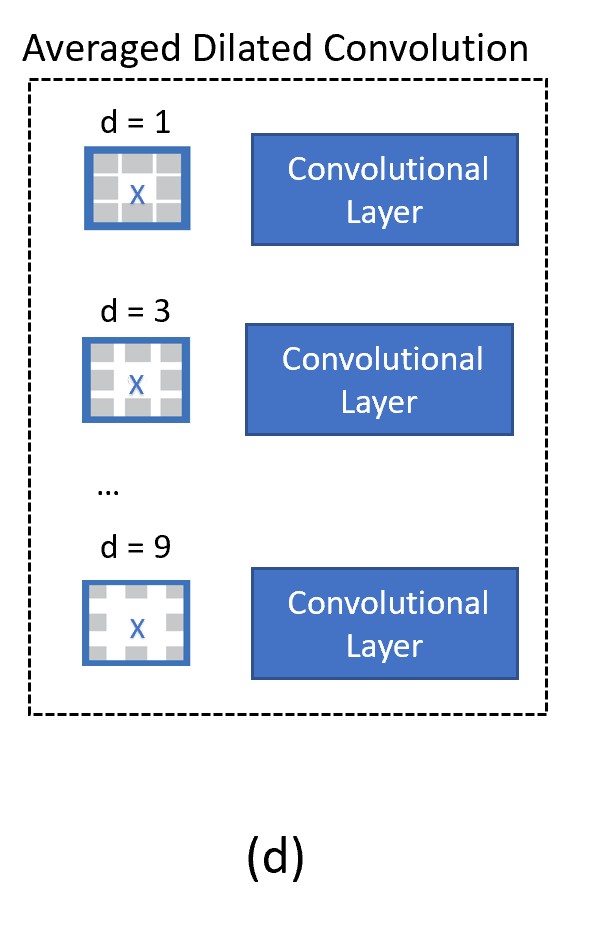}
    \vspace{-10pt}
    \caption{Building blocks of the proposed Architecture: (a) PerceptiveNet architecture, (b) Decoder proposed dilated residual unit (DilRes), (c) Encoder/Bridge proposed dilated residual unit (DilRes), comprised of a mixture of average and maximum pooling layer (Mix Pool) and an averaged Dilated convolutional layer, (d) Dilated convolutional layer. Noticeably, the Mix Pool layer is not present on the decoder.}
    \vspace{-10pt}
    \label{fig:Architecture}
\end{figure*}

Despite advancements in Deep Learning for image processing, tree crown semantic segmentation in dense forest aerial imagery remains challenging. While Convolutional Neural Networks (CNNs) are inherently biased towards learning hierarchies of localised features (Fukushima \cite{fukushima1982neocognitron}), studies have shown they tend to over-rely on texture rather than shape features when trained on standard (Geirhos et al. \cite{geirhos2018imagenet}) or aerial (Voulgaris et al. \cite{voulgaris2023seasonal}) datasets. Relying solely on texture features is particularly problematic in aerial imagery, where texture features vary significantly with lighting, shadows and atmospheric conditions. This exposes a critical gap in current approaches, which often overlook the need for robust shape-based features in aerial scene analysis. Adding fixed Gabor filters helps networks avoid relying on texture, leading to more structured representations (Evans et al. \cite{evans2022biological}). However, this approach restricts the filters to non-adaptive Gabor functions, preventing the network from learning and adapting.

\noindent{\textbf{Contributions.}} This work's contributions are as follows: 1) A novel convolutional layer parameterised by trainable Log-Gabor functions, evaluated both qualitatively and quantitatively against Gabor-based and standard convolutional layers; 2) A new backbone architecture that extracts more salient features, as demonstrated through ablation studies and Class Activation Maps; 3) A novel hybrid CNN-Transformer model leveraging our proposed backbone; and 4) Comprehensive evaluation across multiple aerial datasets and benchmarking against leading models, with extensive qualitative and quantitative analyses demonstrating superior performance in tree crown segmentation and strong generalisation to diverse aerial scene segmentation tasks.

\noindent{\textbf{ Theoretical Insights.}}
Gabor filters combine sinusoidal waves with Gaussian envelopes (Gabor \cite{gabor1946theory}; Daugman \cite{daugman1985uncertainty}), enabling simultaneous spatial and frequency domain analysis. This property makes them particularly suitable for texture analysis and feature extraction in complex imagery:

\noindent{\textit{Real:}}  $g(x, y, \omega, \theta, \psi, \sigma, \gamma) =$ \vspace{-.5em} 
\begin{align}
     \exp\left(- \frac{x'^2 + \gamma^2y'^2}{2\sigma^2}\right) \cos(\omega x' + \psi)
\end{align}

\noindent{\textit{Imaginary:}} $g(x, y, \omega, \theta, \psi, \sigma, \gamma) =$ \vspace{-.5em} 
\begin{align}
         \text{exp}(- \frac{x'^2 + \gamma^2y'^2}{2\sigma^2}) \text{sin}(\omega x' + \psi) \\
                x' = x\text{cos}\theta + y\text{sin}\theta;
        \quad
        y' = -x\text{sin}\theta + y\text{cos}\theta \nonumber
\end{align}

While Gabor filters offer differentiable shapes (Nava et al. \cite{nava2012texture} and Boukerroui et al. \cite{boukerroui2004choice}), they suffer from several limitations:

\begin{itemize}
    \setlength{\leftskip}{1em} 
    \item \textit{Non-Zero DC Component:} The cosine component's non-zero integration limits sensitivity to high-frequency components \cite{heitger1992simulation}, \cite{ronse1993idempotence}.
    \item \textit{Non-Orthogonality:} Overlapping frequency bands across scales reduce discriminative power \cite{granlund2013signal}.
    \item \textit{Non-Uniform Fourier Domain Coverage:} Uneven frequency distribution leads to inadequate high-frequency coverage \cite{boukerroui2004choice}, \cite{fischer2007self}.
    \item \textit{Lack of True Quadrature:} Inconsistent phase differences affect precision in phase-sensitive tasks \cite{fischer2007self}.
\end{itemize}

\noindent{Log-Gabor} filters address these limitations through logarithmic frequency scaling (Fischer et al. \cite{fischer2007self}):
\begin{align}
        G_{\text{log-gabor}}(f) = \exp \left( - \frac{(\log(f / f_0))^2}{2 (\log(\sigma / f_0))^2} \right)
\end{align}
This design provides zero DC components, improved orthogonality across scales, uniform Fourier domain coverage, and superior spatial localisation.

Convolutional filters in the first layer of CNNs often resemble Gabor filters, capturing basic patterns like edges and textures (Luan et al. \cite{luan2018gabor}; Alekseev and Bobe \cite{alekseev2019gabornet}; Evans et al. \cite{evans2022biological}). Both types of filters perform spatial and frequency domain analysis, enhancing the architecture's ability to capture spatial localisation, orientation, and spatial frequency selectivity. This is particularly beneficial for semantic segmentation of tree crowns in dense forests from aerial images, as it improves the network's ability to localise tree crowns, detect various orientations, and differentiate between textures, leading to more accurate and robust segmentation results. Moreover, Log-Gabor filters, with their zero DC components, improved orthogonality, uniform Fourier domain coverage, and superior spatial localisation, further enhance these capabilities, resulting in even more precise and reliable segmentation of tree crowns. In the next sections, we perform a comparative analysis, both quantitatively and qualitatively, to evaluate how each implementation capitalises on the theoretical background and impacts semantic segmentation performance.
\section{Methods}
\label{sec:methods}

This section, introduces the proposed model PerceptiveNet and the hybrid CNN-Transformer PerceptiveNeTr, followed by the test datasets. As PerceptiveNet is based on a ResUNet, we first describe it, and then discuss the changes we have made. We end with a brief description of the models we test in which we isolate each of the changes we have made.

\subsection{Models}
\label{sec:models}
\noindent\textbf{ResUNet.}
This architecture (Zhang et al. \cite{zhang2018road}) extends the U-Net model by incorporating residual units (He et al. \cite{he2016deep}). It is comprised of an encoder and a decoder. The encoder feature maps low-level fine-grained information, whilst the decoder feature maps high-level, coarse-grained semantic information. Skip-connections between low- and high-level feature maps enhance semantic extraction within the encoder and decoder framework.

\noindent\textbf{PerceptiveNet.} Dilated convolution was first introduced by Chen et al. \cite{chen2014semantic} \cite{chen2017deeplab} as a way of increasing the receptive field for the task of semantic segmentation. According to Wei et al. \cite{wei2018revisiting}, convolutional kernel receptive fields are enlarged when employing varying dilation rates, which results in transferring the surrounding discriminative information to the discriminative scene regions. In this work, we propose a novel convolutional layer parameterised by trainable Log-Gabor functions and explore how it performs when combined with averaged dilated convolutions and a mixture of maximum and average pooling layers to impact semantic segmentation performance. Specifically, for the Log-Gabor-parameterised convolutional layer, we used:
\begin{equation}
g(x, y) = g_r(r) \cdot g_\theta(\theta) \cdot \cos(2\pi f_0 r + \psi) \cdot \frac{1}{2\pi\sigma^2}
\end{equation}

\noindent where the radial and angular components are defined as:
\begin{align}
g_r(r) &= \exp\left(-\frac{(\log (r /f_0))^2}{2(\log(\sigma/f_0))^2}\right) \\
g_\theta(\theta) &= \exp\left(-\frac{(\theta - \theta_0)^2}{2\sigma^2}\right)
\end{align}

\noindent and the variables are defined as:
\begin{gather*}
    r = \sqrt{x'^2 + y'^2 + \delta}
\end{gather*}
\vspace{-2.5em}
\begin{align}
    x' &= x\cos\theta + y\sin\theta; &
    y' &= -x\sin\theta + y\cos\theta \nonumber
\end{align}

\noindent Here, $(x, y)$ represents the spatial position, $f_0$ is the centre frequency, $\theta$ is the orientation, $\theta_0$ is the reference orientation, $\sigma$ is the bandwidth parameter, $\psi$ is the phase offset, and $\delta$ is a small constant to prevent division by zero.

Log-Gabor layer weights were initialised by setting the bandwidth parameter $\sigma$, centre frequency $f_0$, and frequency $f$ for each filter. The phase offset $\psi$ is set by uniform distribution $\text{Unif.}(0, \pi)$. Notably, the Log-Gabor function parameters ($f$, $\theta$, $\sigma$, $\psi$, $f_0$, $\theta_0$) are learnable and updated during backpropagation as part of the model’s optimisation.


This implementation offers several advantages over the standard Gabor filter. The logarithmic nature of the radial component allows for a more even coverage of spatial frequencies and can be designed with arbitrary bandwidth. Additionally, Log-Gabor functions have no DC component, which can be beneficial in certain image processing tasks.
The key differences from a standard Gabor filter include:

\begin{itemize}
    \setlength{\leftskip}{1em}
    \item The use of a logarithmic term in the radial component, which provides better frequency coverage.
    \item A normalisation factor of $\frac{1}{2\pi\sigma^2}$, which ensures proper scaling of the filter.
\end{itemize}

\noindent{These} modifications allow the Log-Gabor filter to capture a wider range of spatial frequencies and orientations, potentially improving its performance in tasks such as edge detection, texture analysis, and feature extraction for semantic segmentation.

In addition, due to the information complexity in aerial images, a method that combines maximum and average pooling was applied:
\begin{equation}
    \label{eq:MixPool}
    f_{\text{mix}}(x) = \alpha_l \cdot f_{\text{max}}(x) + (1 - \alpha_l) \cdot f_{\text{avg}}(x)
\end{equation}
\noindent where scalar mixing portion $\alpha_l\in[0,1]$ indicates the max and average combination per layer \textit{l}. For the purpose of this work, we chose a scalar mixing portion $\alpha_l = 0.8$.

The use of pooling layers reduces the spatial resolution of the feature map, while it increases the receptive field of feature points. Thus, each feature point in the feature map is influenced by a larger portion of the input image. By increasing the receptive field, the network can capture larger contextual information, including spatial relationships and dependencies between objects, and this benefits the semantic segmentation tasks where objects may appear at different scales and positions within the image.

Similar to \cite{wei2018revisiting}, an averaged dilated convolutional layer was added on the last convolutional layer of the residual block. Specifically, convolutional blocks with multiple dilated rates (i.e. d = 1, 3, 6, 9) were appended to the final convolutional layer, thus localising scene-related regions observed by different receptive fields. Using high dilation rates (i.e. d = 9) can cause inaccuracies by mistakenly highlighting scene-irrelevant regions. To avoid such scenarios, we used equation \ref{eq:Dil}, where the average over the localisation maps $H_i$ (i.e i = 3, 6, 9) generated by different dilated convolutional blocks was summed to the localisation map $H_0$ of the convolutional block with dilation d = 1.

\begin{equation}
    \label{eq:Dil}
    H = H_o + \frac{1}{n_d} \sum_{n=1}^{n_d}H_i
\end{equation}

Figure \ref{fig:Architecture} illustrates the architecture under consideration. We replaced the initial ResUNet convolutional layer with a convolutional layer parameterised by trainable Log-Gabor functions. Additionally, we propose a residual block for the encoding and bridge parts of the network, where we replaced stride 2 convolutional layers with stride 1. To maintain effective downsampling, we added a mixture of Maximum and Average pooling layers to halve the feature map size, along with an averaged dilated convolutional layer for enhanced feature extraction. For decoding, we employed a residual block with an averaged dilated convolutional layer.

\noindent\textbf{LGMPResUNet.}
To enhance feature extraction, 1) we replaced the first convolutional with a Logarithmic Gabor-parameterised convolutional layer; 2) the encoding path is comprised of three residual blocks. Instead of using the first convolutional layer stride 2 to downsample the feature map size, we used stride 1 and added a mixture of maximum and average pooling layers to reduce the feature map by half.

\noindent\textbf{DilResUNet.}
This model employs dilated convolutional blocks to enhance the representation capacity of the convolution layers when compared to the ResUNet. Thus, added dilated convolutional blocks with multiple dilated rates (i.e. d = 1, 3, 6, 9). This enables the extraction of fine-grained detailed and coarse-grained semantic information. 

\begin{figure}[t!]
    \centering
    \includegraphics[width=1\columnwidth]{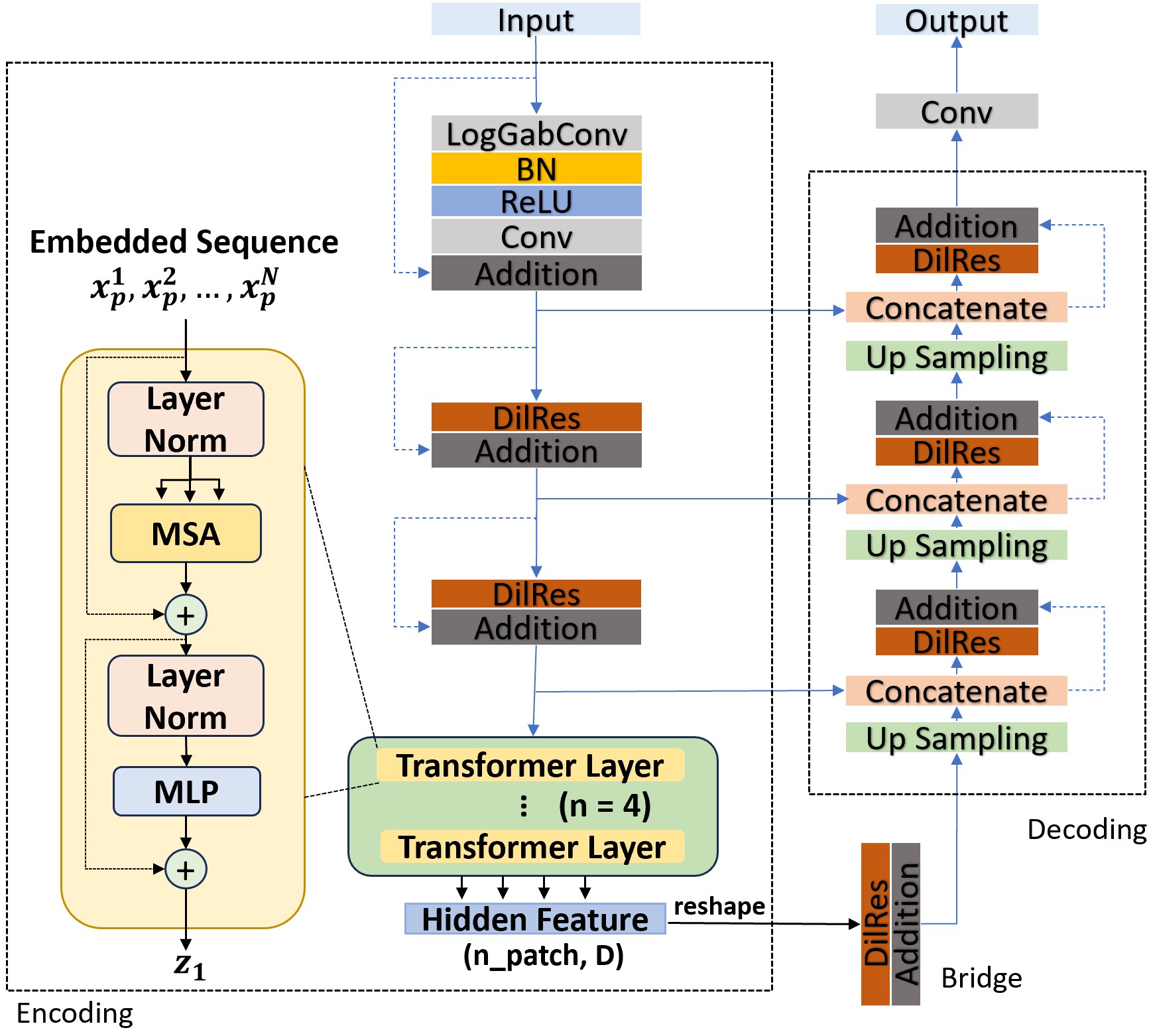}
    \caption{PerceptiveNeTr: Hybrid CNN (PerceptiveNet) - Transformer architecture, leveraging long-range dependencies and global context.}
    \vspace{-10pt}
    \label{fig:Transformer}
\end{figure}

\noindent\textbf{PerceptiveNeTr.} We propose an architecture that combines elements from CNNs and transformer models, enabling the capture of long-range dependencies and global context \cite{dosovitskiy2020vit}. It is structured into three main components: Encoding, Decoding, and a Bridge (Figure \ref{fig:Transformer}). The Encoding section begins the proposed Log-Gabor parameterised convolutional layer, followed by the proposed backbone described above. An embedded sequence processing component is included, representing the Patch Embedding, and features Layer Normalisation, Multi-Head Self-Attention (MSA), and Multi-Layer Perceptron (MLP) layers. Finally, the encoding section consists of a stack of four Transformer layers (n=4), which correspond to the Transformer Encoder described in the model. The Bridge and Decoding section remain as described previously.

\noindent\textbf{ViTResUNet.} This model is the same as the hybrid CNN-Transformer PerceptiveNeTr described above, but uses a ResUNet as the backbone instead.

\noindent \textbf{Model Training.} Due to novelty, all models were trained from scratch for 130 epochs (based on when train/validation results stopped improving), using Adam Optimiser and Cross Entropy loss and a batch size of 16. The data was split as 80\% train/validation (of which, 80\% train; 20\% validation) and 20\% for testing. The only data augmentations applied during training were geometric transformations, i.e rotation with 90\% probability and horizontal and vertical flip with 50\% and 10\% probabilities respectively.

\noindent\textbf{Performance Metrics.} To measure model performance two scores are used, Pixel accuracy (Acc) and mean Intersection of a Union (mIoU) score. As pixel accuracy (calculated as the proportion of correctly predicted pixels) might not accurately reflect a model's performance when e.g. water or ground scenes dominate an image, we additionally use mIoU as it takes into account the area covered by each of the k classes via:
\begin{equation}
    \text{mIoU} = \frac{1}{k + 1} \sum_{i=0}^{k} \frac{\text{Area of Overlap} \includegraphics[keepaspectratio = true, scale = .14]{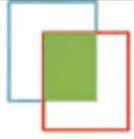}}{\text{Area of Union} \includegraphics[keepaspectratio = true, scale = .14]{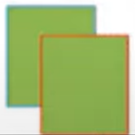}}
    \vspace{-5pt}
\end{equation}

\subsection{Data}
\label{sec:Data}
Alongside TreeCrown, two additional datasets were chosen to evaluate model generalisation across diverse aerial scenes of varying complexity. While differing from dense forests, both exhibit similar challenges, such as complex spatial boundaries and occlusions caused by canopy and shadows.

\noindent\textbf{TreeCrown.}
The dataset by Cloutier et al. \cite{cloutier2024influence} covers a temperate-mixed forest in the Laurentides region of Québec, Canada, during 2021. It includes 1,360 (768$\times$768-pixels) UAV images acquired monthly from May to August. The dataset contains 23,000 segmented tree crowns with per-pixel species annotations for 14 tree species classes.



\noindent\textbf{Landcover.AI.}
The dataset by Boguszewski et al. \cite{Boguszewski_2021_CVPR} depicts RGB aerial images and annotated buildings, forests, water bodies and roads in Poland. The dataset is comprised of 33 orthophotos with 25 cm per pixel resolution (9000$\times$9500-pixels) and 8 orthophotos with 50 cm per pixel resolution (4200$\times$4700-pixels) of a total area of 216.27 km$^2$. We sub-sampled the orthophotos to produce 21,924 (256$\times$256-pixels) images and their corresponding masks.

\noindent\textbf{UAVid.}
This is a UAV semantic segmentation dataset by Lyu et al. \cite{lyu2020uavid}, that focuses on urban scenes with two spatial resolutions (3840$\times$2160 and 4096$\times$2160-pixels) and eight class labels (Building, Road, Static car, Tree, Low vegetation, Human, Moving car, and Background). Each image was padded and cropped into eight 256$\times$256-pixel patches.
\section{Results and Discussion}
\label{sec:Results}

\subsection{Comparative Analysis: Standard, Gabor and Log-Gabor Convolutional Layers}

In this section, we evaluate the impact of using a standard initial convolutional layer, a Gabor-based, and a Log-Gabor parameterised initial convolutional layer in tree crown semantic segmentation. The theoretical advantages of Log-Gabor filters—including improved frequency coverage, zero DC component, reduced artefacts, enhanced edge detection, and adaptability to shape variability—suggest that the Log-Gabor parameterised convolutional layer is likely to extract more salient features compared to both the Gabor-based and standard convolutional layers. By conducting qualitative and quantitative analyses, we investigate how these theoretical insights are reflected in the experimental results, thereby illustrating the effectiveness of our proposed Log-Gabor parameterised convolutional layer in accurately segmenting tree crowns in dense forests.

\noindent\textbf{Quantitative Analysis.}
We evaluate the impact of using a standard, Gabor-based, and trainable Log-Gabor function-parameterised initial convolutional layer within our proposed backbone on semantic segmentation performance. Table \ref{tab:GabLogGabResults} shows that in the complex TreeCrown dataset, the parameterised Log-Gabor outperforms both the convolution and the Gabor one by $4.6\%$ and $3.7\%$ mIoU respectively. Experiments on Landcover.AI and UAVid datasets confirm this trend, with Log-Gabor outperforming conventional and Gabor implementations by $3.5\% $ and $2.8\%$ mIoU on Landcover.AI, and $4.5\%$ and $1.5\%$ mIoU on UAVid, respectively.

\begin{table}[!h]
    \raggedright
    \vspace{-6pt}
    \caption{Convolutional vs Gabor vs Log-Gabor Parameterised Initial Convolutional Layer Semantic Segmentation Performances.}
    \label{tab:GabLogGabResults}
     \setlength{\tabcolsep}{2.1pt} 
    \scalebox{0.8}{
    \begin{tabular*}{\columnwidth}{@{\extracolsep{\fill}} l|cc|cc|cc}
        \cmidrule{1-7}
        \textbf{Dataset} & \multicolumn{2}{c|}{TreeCrown} & \multicolumn{2}{c|}{Landcover.AI} & \multicolumn{2}{c}{UAVid} \\
        \textbf{Architecture} & \small\textbf{Acc(\%)} & \small\textbf{mIoU(\%)} & \small\textbf{Acc(\%)} & \small\textbf{mIoU(\%)} & \small\textbf{Acc(\%)} & \small\textbf{mIoU(\%)} \\
        \cmidrule{1-7}
        \small{PerceptiveNet}$^{\text{Conv}}$ & 82.9 & 43.5 & 91.3 & 78.1 & 85.3 & 64.1 \\
        \small{PerceptiveNet}$^{\text{Gab}}$ & 82.2 & 44.4 & 92.1 & 78.8 & 86.2 & 67.1 \\
        \footnotesize{PerceptiveNet}$^{\text{LogGab}}$ & \textbf{84.4} & \textbf{48.1} & \textbf{92.5} & \textbf{81.6} & \textbf{87.1} & \textbf{68.6} \\
        \cmidrule{1-7}
    \end{tabular*}
    }
    \vspace{-8pt}
\end{table}

\noindent\textbf{Qualitative Analysis.} 
We perform a visual inspection of the segmentation masks predicted when using the two initial Log-Gabor and Gabor parameterised convolutional layers under review. As is evident in Figure \ref{fig:QualitativeTreeCrown}, incorporating a Log-Gabor convolutional layer allows the model to capture finer spatial details, resulting in more accurate segmentation across different tree crown sizes. Moreover, the Log-Gabor convolutional layer is very efficient at differentiating closely packed trees, effectively managing complex crown shapes and smaller, irregular crowns. Particularly in the presence of dead trees, Log-Gabor outperforms Gabor or normal convolutional layers by capturing more high-frequency features, enhancing its overall segmentation performance. This demonstrates the importance of zero DC components, improved orthogonality across scales, uniform Fourier domain coverage, and spatial localisation. These properties enhance the model's ability to perform semantic segmentation tasks in complex, challenging dense forests.

\subsection{Ablation Study}

As the proposed architecture comprises of two components, a texture-biased part for extracting salient features and a wider receptive field, we evaluated how each of these components contributes to segmentation performance. 

\begin{table}[!h]
    \centering
    \vspace{-6pt}
    \caption{Ablation Study - Semantic Segmentation Performance.}
    \label{tab:AblationResults}
    \setlength{\tabcolsep}{1.3pt} 
    \scalebox{0.86}{
    \begin{tabularx}{\textwidth}{p{2cm}|cc|cc|cc}
        \cmidrule[0.8pt]{1-7}
        \textbf{Dataset} & \multicolumn{2}{c|}{TreeCrown} & \multicolumn{2}{c|}{Landcover.AI} & \multicolumn{2}{c}{UAVid} \\
        \textbf{Architecture} & \small \textbf{Acc(\%)} & \small \textbf{mIoU(\%)} & \small \textbf{Acc(\%)} & \small \textbf{mIoU(\%)} & \small \textbf{Acc(\%)} & \small \textbf{mIoU(\%)} \\
        \cmidrule{1-7}
        ResUNet & 79.3 & 37.6 & 91.3 & 72.6 & 82.6 & 59.7 \\ 
        DilResUNet & 80.6 & 40.5 & 89.4 & 76.8 & 82.9 & 63.1 \\
        \small{LGMPResUNet} & 82.2 & 44.4 & 90.6 & 78.4 & 85.8 & 66.4 \\
        PerceptiveNet & \textbf{84.4} & \textbf{48.1} & \textbf{92.5} & \textbf{81.6} & \textbf{87.1} & \textbf{68.6} \\
        \cmidrule[0.8pt]{1-7}
    \end{tabularx}
    }
    \vspace{-8pt}
\end{table}

Table \ref{tab:AblationResults} results indicate that DilResUNet architecture, with averaged dilated convolutional residual blocks, improves mIoU by $2.9\%$ (TreeCrown), $4.2\%$ (Landcover.AI), and $3.4\%$ (UAVid) over ResUNet. LGMPResUNet, integrating an initial Log-Gabor parameterised convolutional layer and residual blocks comprised of a mixture of average and maximum pooling layers, further boosts mIoU by $6.8\%$, $5.8\%$, and $6.7\%$, respectively. The proposed PerceptiveNet, combining an initial Log-Gabor parameterised convolutional layer, residual blocks consisting of a mixture of average and maximum pooling layers, and averaged dilated convolutional layer, achieves the best mIoU scores of $10.5\%$ (TreeCrown), $9.0\%$ (Landcover.AI), and $8.9\%$ (UAVid) compared to ResUNet. This study demonstrates that the proposed layers complement each other to further enhance the semantic segmentation performance.

\subsection{Comparative Feature Extraction Analysis}

\begin{figure}[!htbp]
    \begin{tabularx}{\columnwidth}{
        *{3}{>{\centering\arraybackslash}X} 
    }
        \toprule
        \textbf{\small Image} &
        \textbf{\small ResUNet} &
        \textbf{\small PerceptiveNet} \\
        \midrule
        \begin{subfigure}{\linewidth}
            \centering
            \includegraphics[width=.9\linewidth]{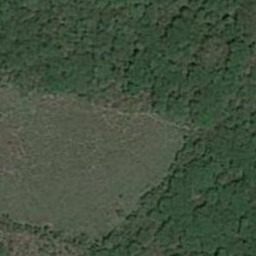}
        \end{subfigure} &
        \begin{subfigure}{\linewidth}
            \centering
            \includegraphics[width=.9\linewidth]{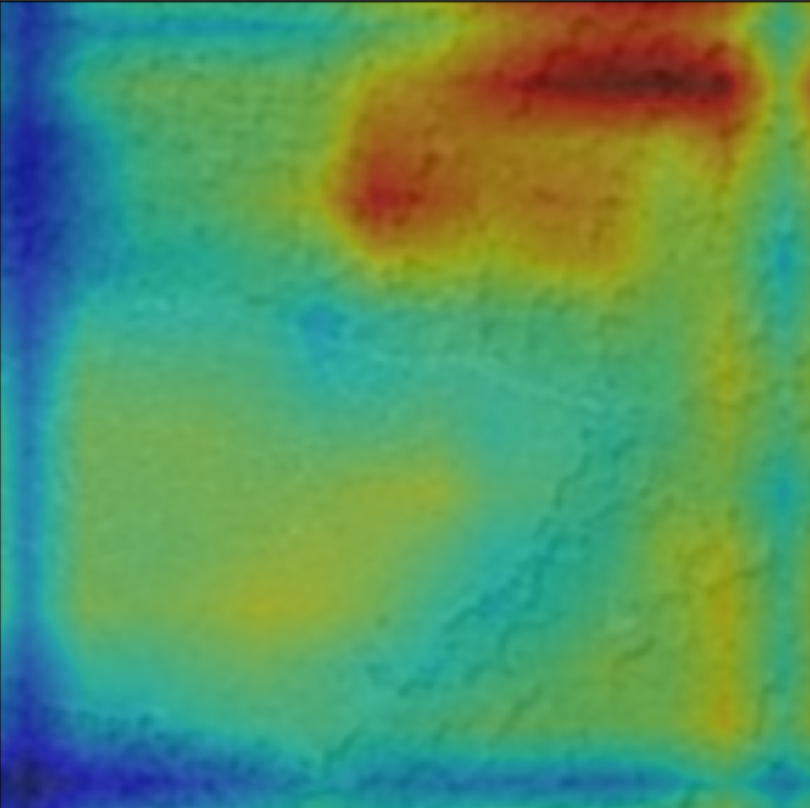}
        \end{subfigure} &
        \begin{subfigure}{\linewidth}
            \centering
            \includegraphics[width=.9\linewidth]{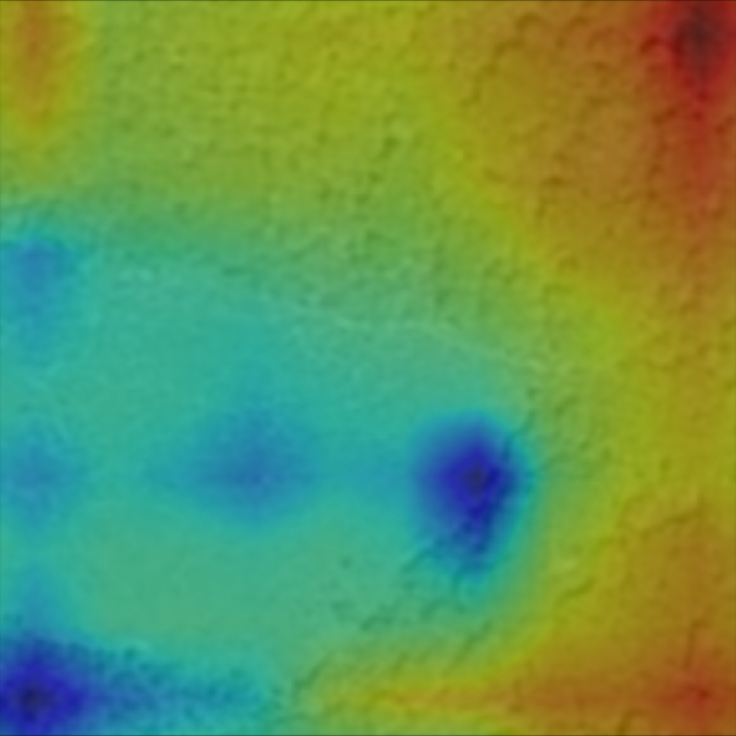}
        \end{subfigure} \\
        \begin{subfigure}{\linewidth}
            \centering
            \includegraphics[width=.9\linewidth]{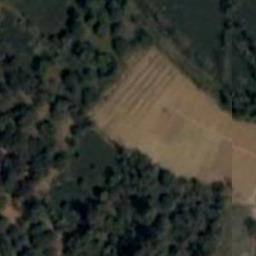}
        \end{subfigure} &
        \begin{subfigure}{\linewidth}
            \centering
            \includegraphics[width=.9\linewidth]{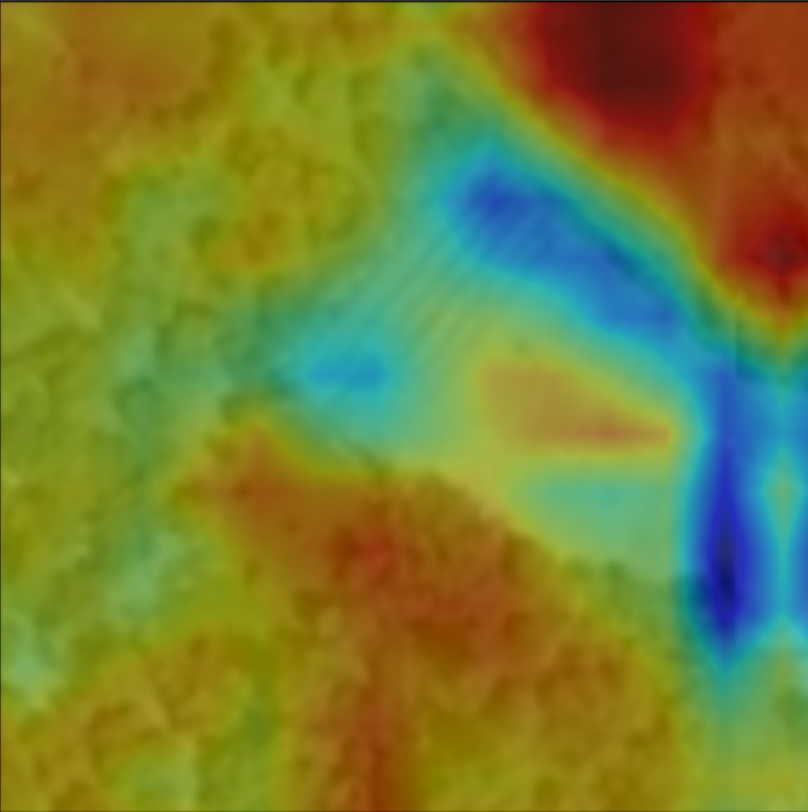}
        \end{subfigure} &
        \begin{subfigure}{\linewidth}
            \centering
            \includegraphics[width=.9\linewidth]{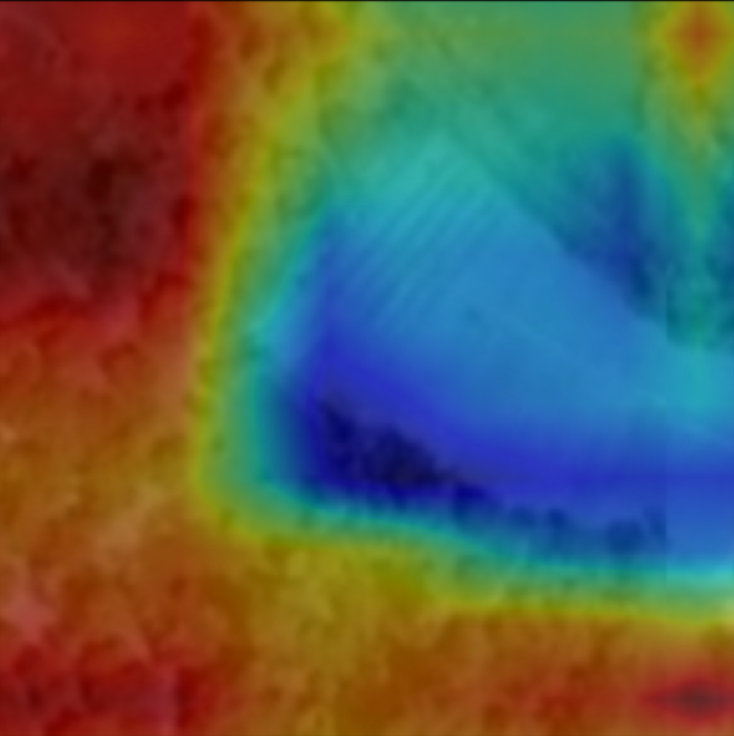}
        \end{subfigure} \\
        \begin{subfigure}{\linewidth}
            \centering
            \includegraphics[width=.9\linewidth]{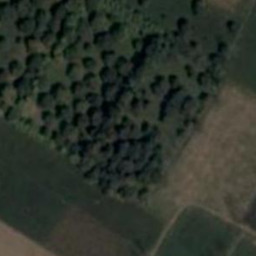}
        \end{subfigure} &
        \begin{subfigure}{\linewidth}
            \centering
            \includegraphics[width=.9\linewidth]{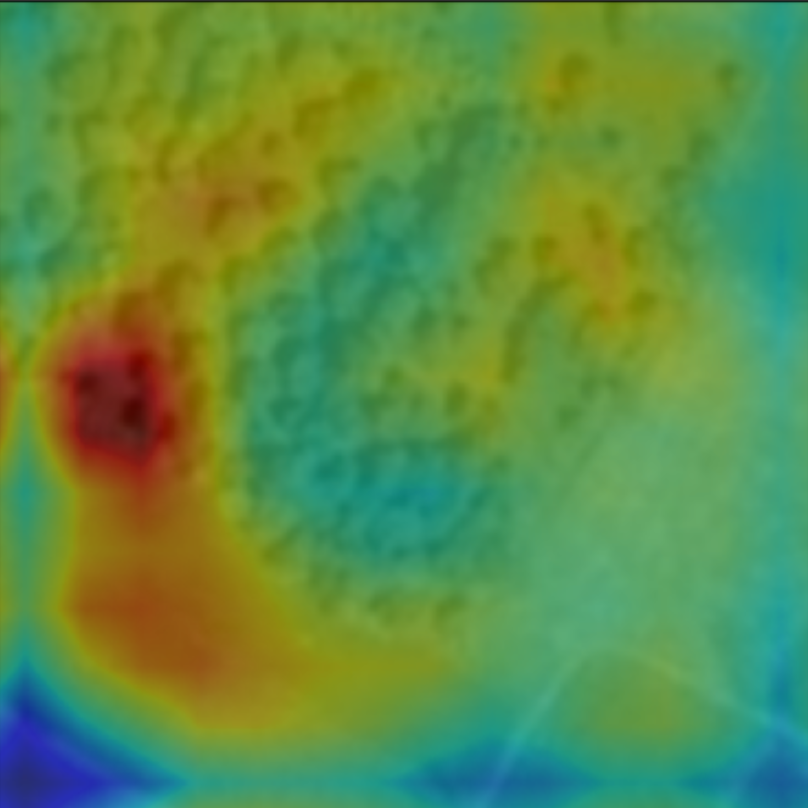}
        \end{subfigure} &
        \begin{subfigure}{\linewidth}
            \centering
            \includegraphics[width=.9\linewidth]{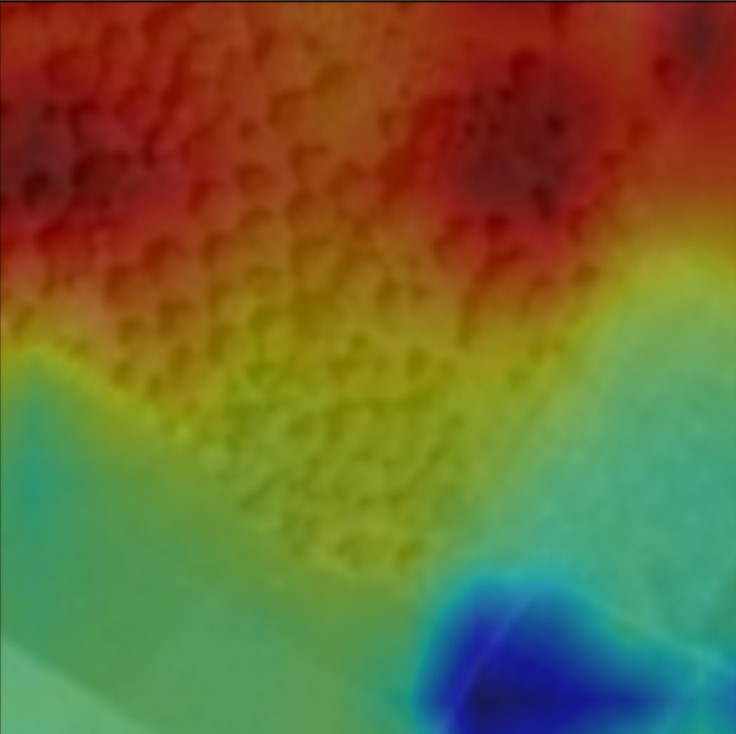}
        \end{subfigure} \\
        \begin{subfigure}{\linewidth}
            \centering
            \includegraphics[width=.9\linewidth]{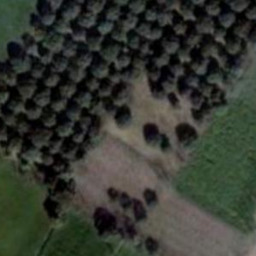}
        \end{subfigure} &
        \begin{subfigure}{\linewidth}
            \centering
            \includegraphics[width=.9\linewidth]{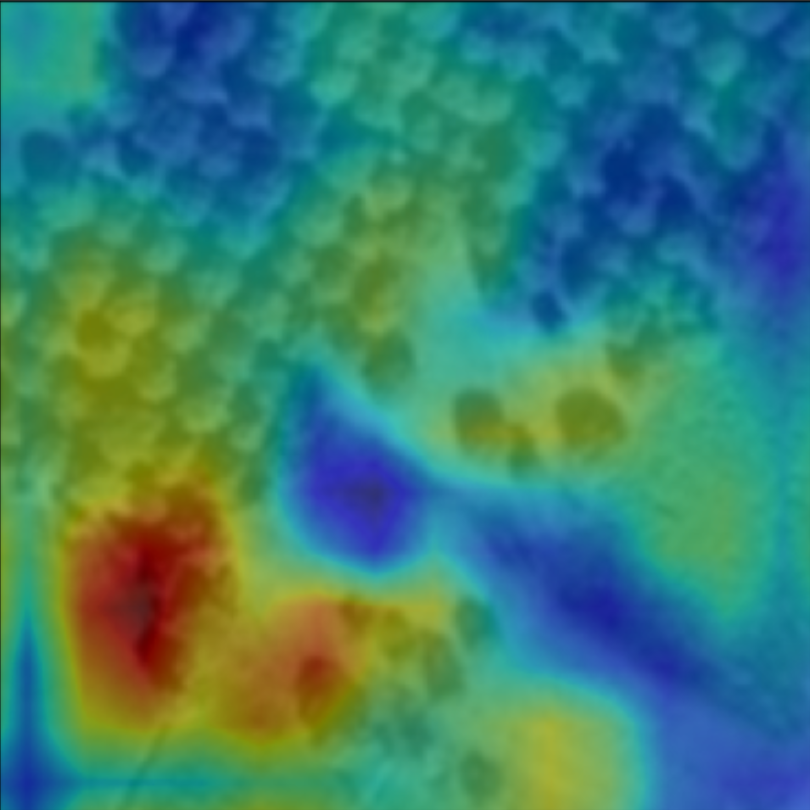}
        \end{subfigure} &
        \begin{subfigure}{\linewidth}
            \centering
            \includegraphics[width=.9\linewidth]{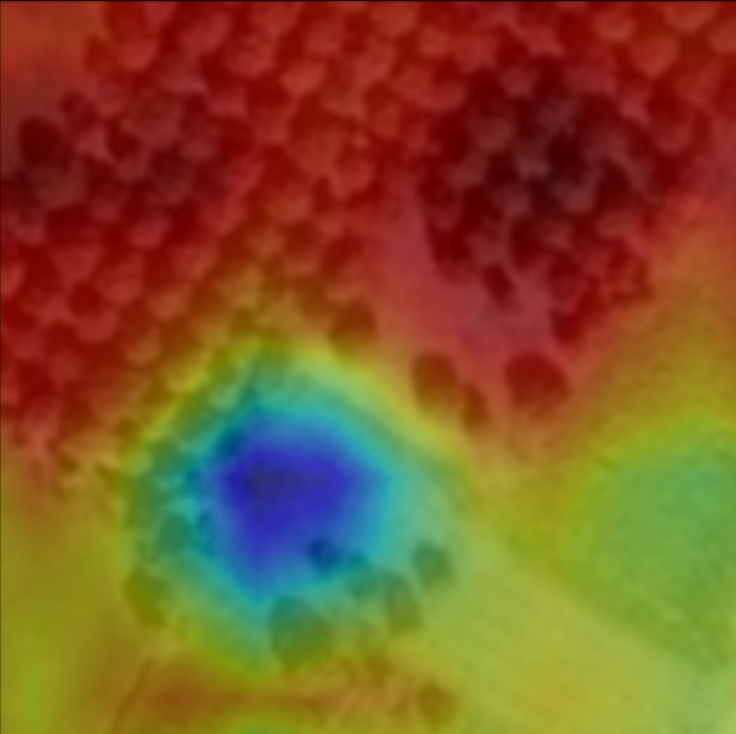}
        \end{subfigure} \\
       \bottomrule
    \end{tabularx}
    \caption{Class Activation Maps, ResUNet vs PerceptiveNet Encoder feature extraction capabilities. PerceptiveNet: More focused and detailed activation pattern.}
    \vspace{-10pt}
    \label{fig:CAM}
\end{figure}

In this section, we employ Class Activation Mapping (CAM, Zhou et al. \cite{zhou2016learning}) to gain insight into the model's decision process by overlaying a heatmap on the original image, indicating the discriminative region used by the model when predicting that an image belongs to a particular class. Figure \ref{fig:CAM} illustrates the CAMs for images containing forests and agricultural land, comparing the feature extraction capabilities of a standard CNN with those of the proposed model. In the ResUNet column, the standard model's activation patterns are diffuse, with focus areas spread across both forested regions and agricultural lands. This scattered attention suggests inefficient feature capture, missing critical details and leading to less robust representations.

In contrast, the PerceptiveNet model demonstrates a more focused and detailed activation pattern, particularly in areas with dense tree coverage. This is expected due to the model's enhanced ability to extract salient features combined with a wider receptive field. Thus, the activation maps display more defined structures and higher contrast, especially in forested regions. This targeted focus indicates that our model develops more robust and specific representations of forest structures. The enhanced ability to concentrate on relevant features, particularly trees, suggests better out-of-distribution generalisation, crucial for tasks like dense forest tree crown segmentation. Comparing these activation maps, we infer that the proposed model's architecture allows for more efficient and targeted feature extraction, potentially improving performance.

\subsection{Quantitative Analysis}

We evaluate the semantic segmentation performance of our proposed architectures through two analyses. First, we compare the PerceptiveNet against State-Of-The-Art (SOTA) CNN models including U-Net \cite{ronneberger2015u}, MACUNet \cite{li2021multiattention}, ResUNet \cite{zhang2018road}, and DeepLabV3+ \cite{chen2018encoder}. Second, we investigate how semantic segmentation performance benefits from the combination of capturing long-range dependencies and global context through our hybrid CNN-Transformer model (ViTResUNet), comparing it against SOTA Transformer architectures UNETR\_2D \cite{hatamizadeh2022unetr} and SwinUNet \cite{cao2022swin}, as well as examine the generalisation of the proposed backbone (PerceptiveNeTr) within such a hybrid framework. All evaluations are conducted across three complex aerial scene SOTA datasets: TreeCrown, Landcover.AI, and UAVid.

\subsubsection{Convolutional Models}

For the TreeCrown dataset in Table \ref{tab:Results}, ResUNet achieved an mIoU of $37.6\%$. U-Net improved this by $5.6\%$. MACUNet further enhanced the results to $45.9\%$, an $8.3\%$ improvement, while DeeplabV3+ showed a similar improvement with an mIoU of $45.7\%$, $8.1\%$ higher than ResUNet. The proposed PerceptiveNet outperformed all models with an mIoU of $48.1\%$, a $10.5\%$ improvement over ResUNet.

\begin{table}[!h]
    \centering
    \vspace{-6pt}
    \caption{Semantic Segmentation Performance in Various Datasets.}
    \label{tab:Results}
    \setlength{\tabcolsep}{1.3pt} 
    \scalebox{0.88}{
    \begin{tabularx}{\textwidth}{p{2cm}|cc|cc|cc}
        \cmidrule[0.8pt]{1-7}
        \textbf{Dataset} & \multicolumn{2}{c|}{TreeCrown} & \multicolumn{2}{c|}{Landcover.AI} & \multicolumn{2}{c}{UAVid} \\
         \textbf{Architecture} & \small \textbf{Acc(\%)} & \small \textbf{mIoU(\%)} & \small \textbf{Acc(\%)} & \small \textbf{mIoU(\%)} & \small \textbf{Acc(\%)} & \small \textbf{mIoU(\%)} \\
        \cmidrule{1-7}
        ResUNet & 79.3 & 37.6 & 91.3 & 72.6 & 82.6 & 59.7 \\ 
        UNet & 81.9 & 43.2 & 91.1 & 73.1 & 85.7 & 64.9 \\
        MACUNet & 83.7 & 45.9 & 91.4 & 74.3 & 85.7 & 66.1 \\
        DeepLabV3+ & 81.9 & 45.7 & 91.9 & 78.6 & 86.7 & 65.8 \\
        PerceptiveNet & \textbf{84.4} & \textbf{48.1} & \textbf{92.5} & \textbf{81.6} & \textbf{87.1} & \textbf{68.6} \\
        \cmidrule[0.8pt]{1-7}
    \end{tabularx}
    }
    \vspace{-8pt}
\end{table}

In the Landcover.AI dataset, ResUNet recorded an mIoU of $72.6\%$. U-Net achieved $73.1\%$ mIoU, $0.5\%$ higher. MACUNet achieved an mIoU of $74.3\%$, which is $1.7\%$ higher. DeepLabV3+, with a mIoU score of $78.6\%$ improved by $6\%$. PerceptiveNet model achieved the best performance, with an mIoU of $81.6\%$, $9\%$ higher than ResUNet.

For the UAVid dataset, ResUNet attained an mIoU of $59.7\%$. The U-Net improved to $64.9\%$ mIoU, $5.2\%$ higher. MACUNet achieved $66.1\%$, $6.4\%$ higher. DeepLabV3+ reached $65.8\%$, a $6.1\%$ improvement. The PerceptiveNet model led with $68.6\%$ mIoU, $8.9\%$ higher than ResUNet.

\subsubsection{Transformer Models}

The summarised results in Table \ref{tab:TansCNNResults} indicate 2 key findings:

\begin{table}[!h]
    \centering
    \vspace{-6pt}
    \caption{Transformer vs Hybrid CNN-Transformer Semantic Segmentation Performance in Various Datasets.}
    \label{tab:TansCNNResults}
    \setlength{\tabcolsep}{1.3pt} 
    \scalebox{0.88}{
    \begin{tabularx}{\textwidth}{p{2cm}|cc|cc|cc}
        \cmidrule[0.8pt]{1-7}
        \textbf{Dataset} & \multicolumn{2}{c|}{TreeCrown} & \multicolumn{2}{c|}{Landcover.AI} & \multicolumn{2}{c}{UAVid} \\
        \textbf{Architecture} & \small\textbf{Acc(\%)} & \small\textbf{mIoU(\%)} & \small\textbf{Acc(\%)} & \small\textbf{mIoU(\%)} & \small\textbf{Acc(\%)} & \small\textbf{mIoU(\%)} \\
        \cmidrule{1-7}
        UNETR\_2D & 75.2 & 27.1 & 81.2 & 52.4 & 80.4 & 55.9 \\
        SwinUNet & 76.9 & 36.5 & 89.1 & 67.3 & 82.3 & 58.3 \\
        ViTResUNet & 79.9 & 37.8 & 89.1 & 68.2 & 83.9 & 60.1 \\
        \small{PerceptiveNeTr} & \textbf{81.8} & \textbf{42.0} & \textbf{91.2} & \textbf{75.3} & \textbf{84.8} & \textbf{64.3} \\
        \cmidrule[0.8pt]{1-7}
    \end{tabularx}
    }
    \vspace{-8pt}
\end{table}

\noindent\textbf{Proposed Hybrid CNN-Transformer:} The ViTResUNet model outperforms the pure Transformer models UNETR\_2D and SwinUNet by $10.7\%$ \& $1.3\%$ mIoU (TreeCrown), $15.8\%$ \& $0.9\%$ mIoU (Landcover.AI), and $4.2\%$ \& $0.8\%$ mIoU (UAVid). This demonstrates that hybridising CNN and Transformer elements benefits from capturing long-range dependencies and global context, leading to enhanced performance in semantic segmentation tasks.

\noindent\textbf{PerceptiveNet Backbone Impact:} The proposed Log-Gabor parameterised convolutional layer and backbone significantly enhance the performance of the hybrid model. PerceptiveNeTr achieves higher mIoU scores compared to ViTResUNet, improving even further the performance by $4.2\%$ mIoU for TreeCrown, $7.1\%$ mIoU for Landcover.AI, and $4.2\%$ mIoU for UAVid datasets. This demonstrates the generalisability of our PerceptiveNet model, indicating the benefits of implementing the proposed backbone even within a hybrid CNN-Transformer framework. 

Overall, the proposed architectures demonstrated superior performance, with PerceptiveNet achieving the highest mIoU scores amongst all models and datasets and PerceptiveNeTr outperforming all Transformer architectures. The integration of Log-Gabor parameterised convolutional layer, residual blocks with mixed pooling layers and averaged dilated convolutions enhanced semantic segmentation in both standalone and hybrid implementations, proving effective at capturing complex features across various models.

\subsection{Qualitative Analysis}

\begin{figure*}[h]
    \begin{tabularx}{\linewidth}{
        *{5}{>{\centering\arraybackslash}X} 
    }
        \toprule
        \textbf{\small Original} &
        \textbf{\small Mask} &
        \textbf{\small ResUNet} &
        \textbf{\small Gabor PerceptiveNet} &
        \textbf{\small LogGab PerceptiveNet} \\
        \midrule
        \begin{subfigure}{\linewidth}
            \centering
            \includegraphics[width=.8\linewidth]{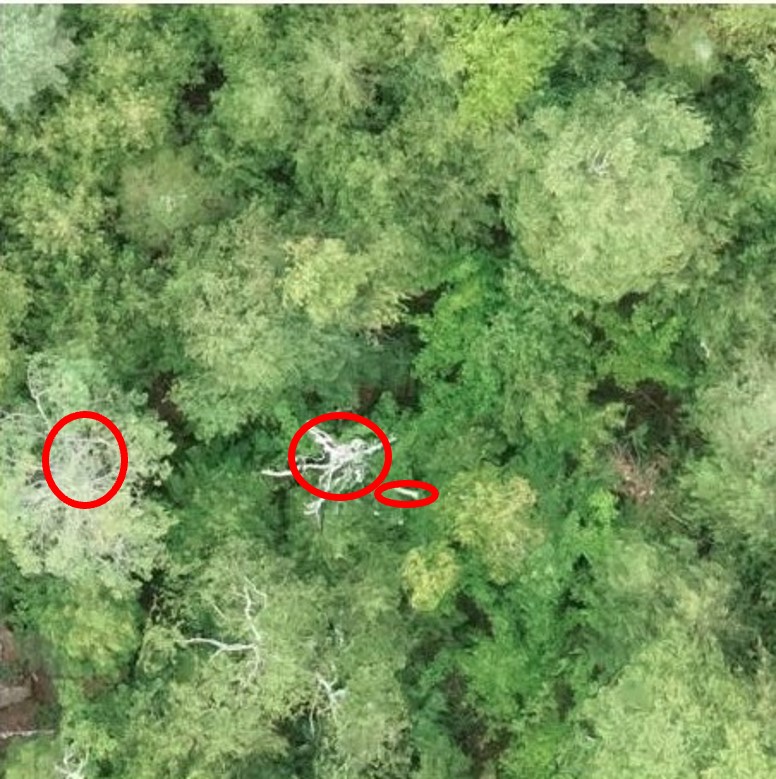}
        \end{subfigure} &
        \begin{subfigure}{\linewidth}
            \centering
            \includegraphics[width=.8\linewidth]{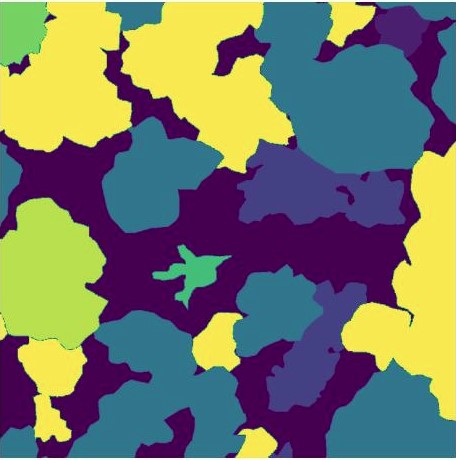}
        \end{subfigure} &
        \begin{subfigure}{\linewidth}
            \centering
            \includegraphics[width=.8\linewidth]{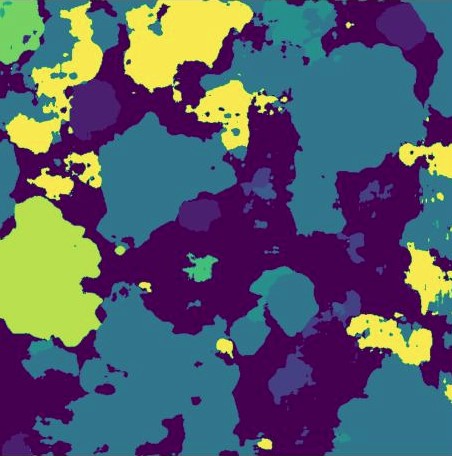}
        \end{subfigure} &
        \begin{subfigure}{\linewidth}
            \centering
            \includegraphics[width=.8\linewidth]{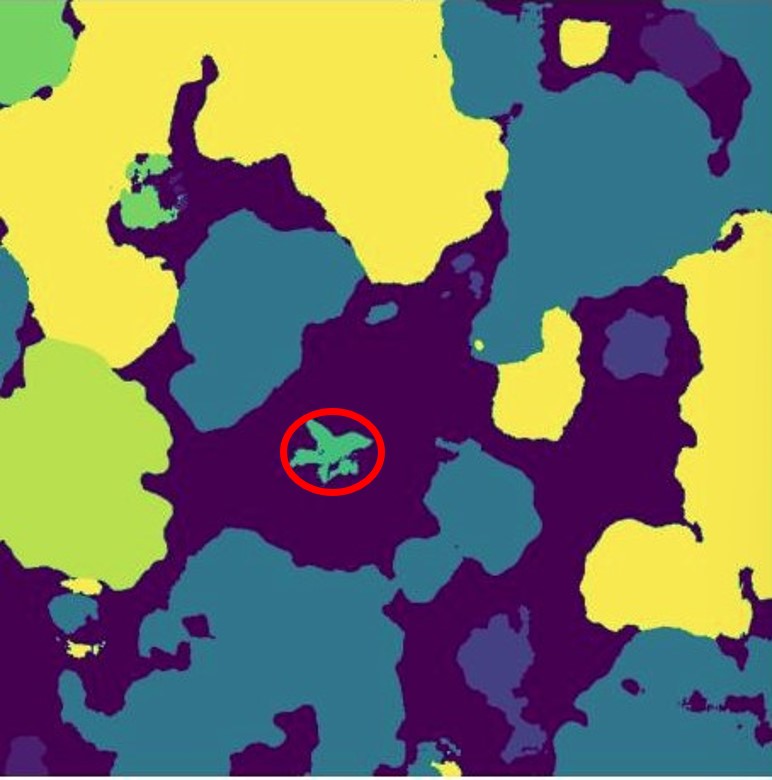}
        \end{subfigure} &
        \begin{subfigure}{\linewidth}
            \centering
            \includegraphics[width=.8\linewidth]{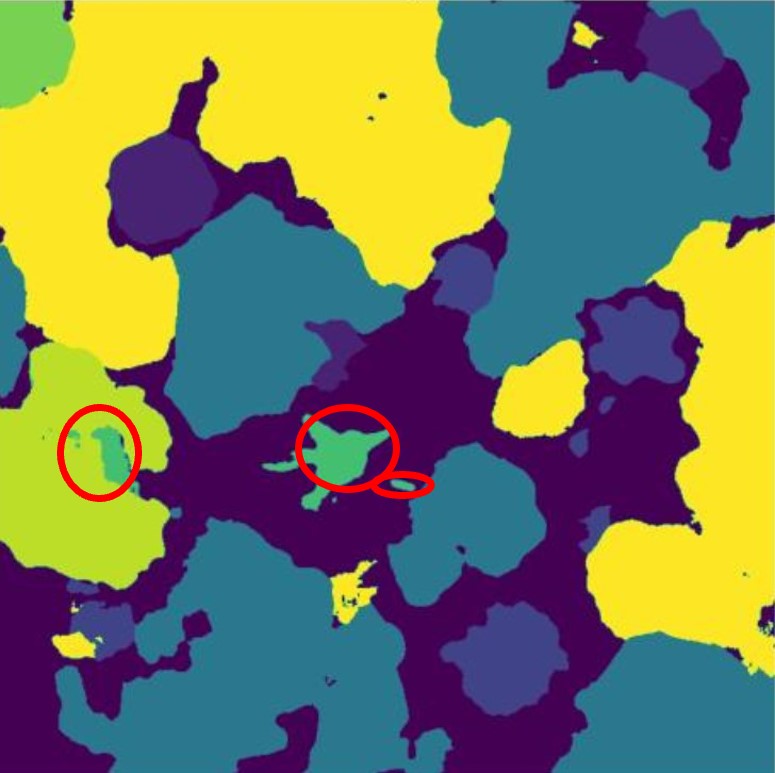}
        \end{subfigure} \\
        \begin{subfigure}{\linewidth}
            \centering
            \includegraphics[width=.8\linewidth]{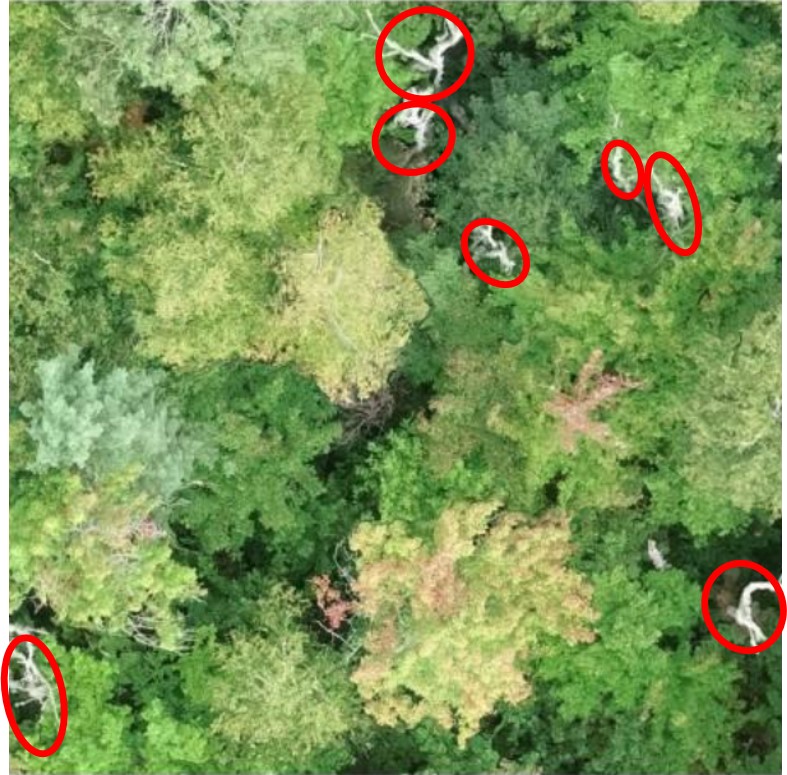}
        \end{subfigure} &
        \begin{subfigure}{\linewidth}
            \centering
            \includegraphics[width=.8\linewidth]{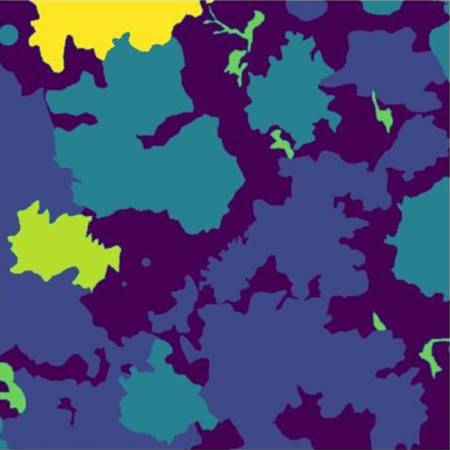}
        \end{subfigure} &
        \begin{subfigure}{\linewidth}
            \centering
            \includegraphics[width=.8\linewidth]{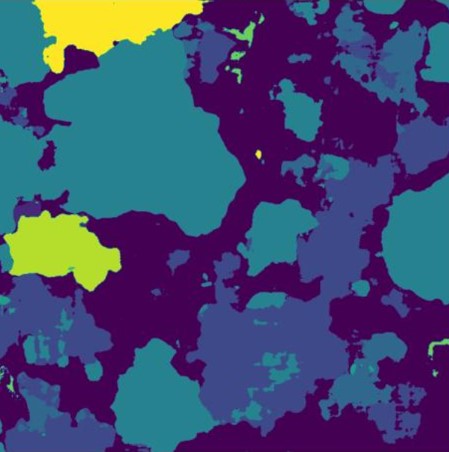}
        \end{subfigure} &
        \begin{subfigure}{\linewidth}
            \centering
            \includegraphics[width=.8\linewidth]{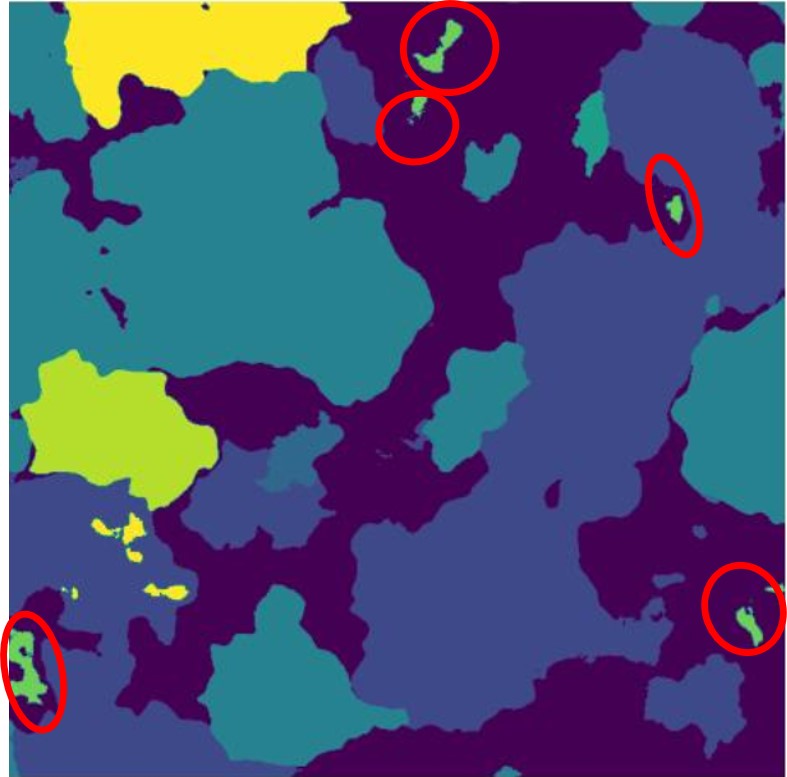}
        \end{subfigure} &
        \begin{subfigure}{\linewidth}
            \centering
            \includegraphics[width=.8\linewidth]{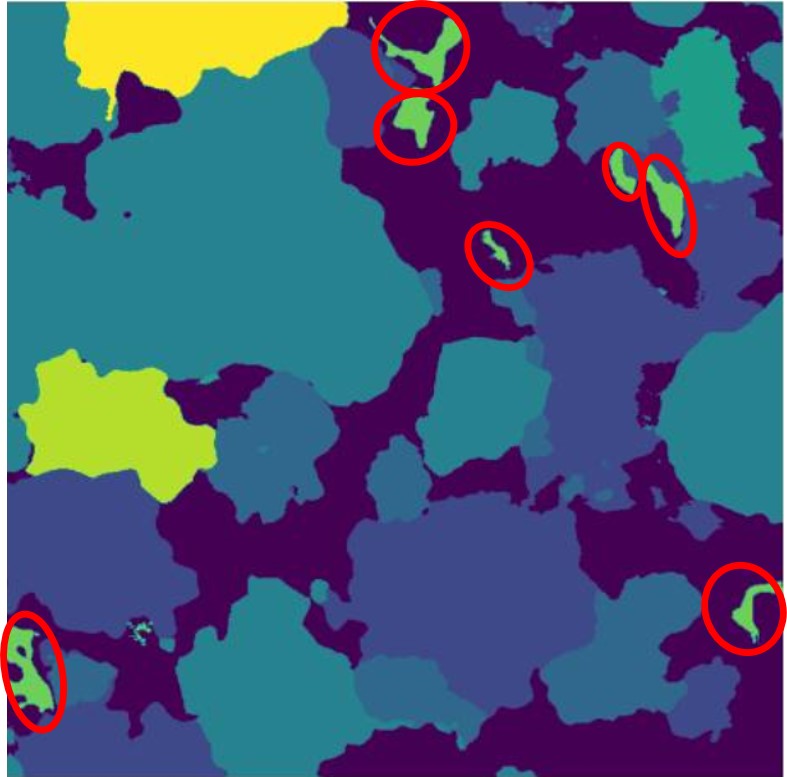}
        \end{subfigure} \\
        \begin{subfigure}{\linewidth}
            \centering
            \includegraphics[width=.8\linewidth]{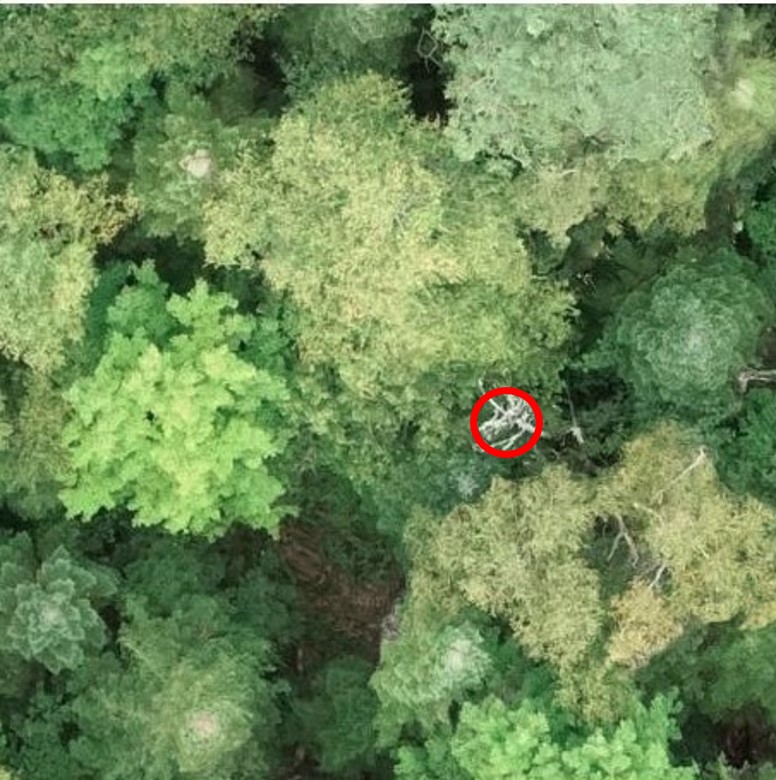}
        \end{subfigure} &
        \begin{subfigure}{\linewidth}
            \centering
            \includegraphics[width=.8\linewidth]{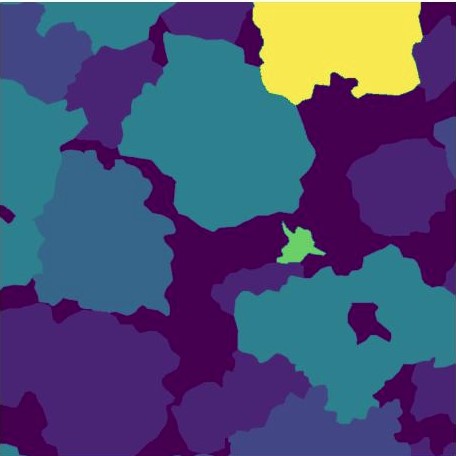}
        \end{subfigure} &
        \begin{subfigure}{\linewidth}
            \centering
            \includegraphics[width=.8\linewidth]{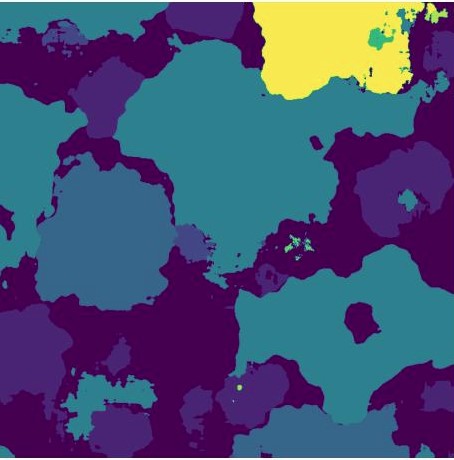}
        \end{subfigure} &
        \begin{subfigure}{\linewidth}
            \centering
            \includegraphics[width=.8\linewidth]{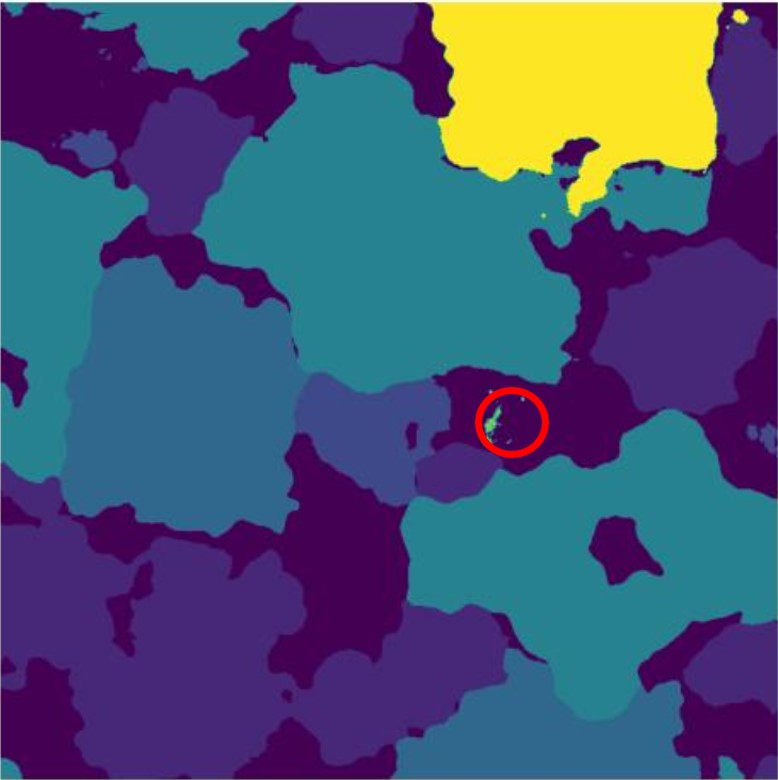}
        \end{subfigure} &
        \begin{subfigure}{\linewidth}
            \centering
            \includegraphics[width=.8\linewidth]{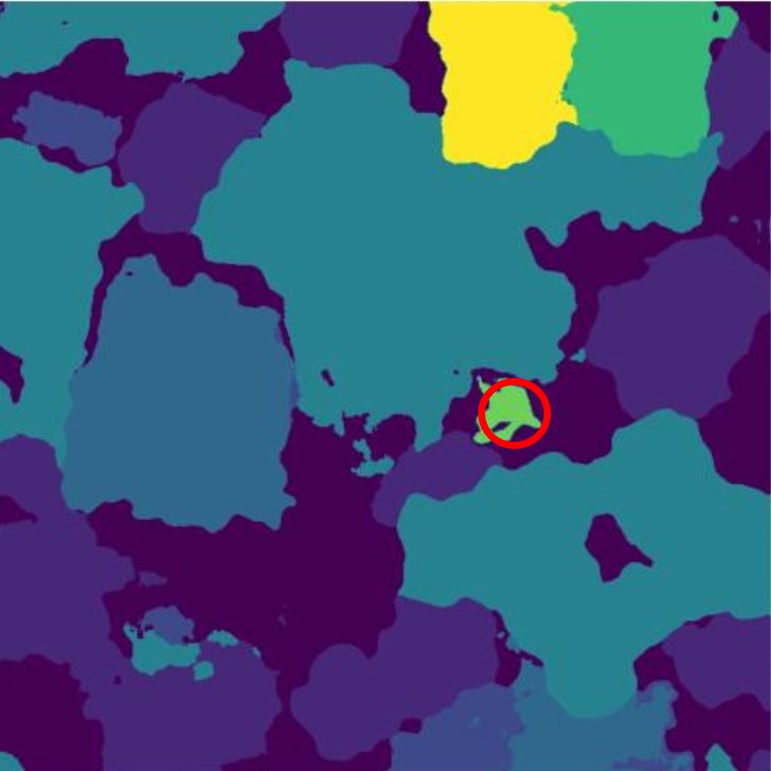}
        \end{subfigure} \\
        \begin{subfigure}{\linewidth}
            \centering
            \includegraphics[width=.8\linewidth]{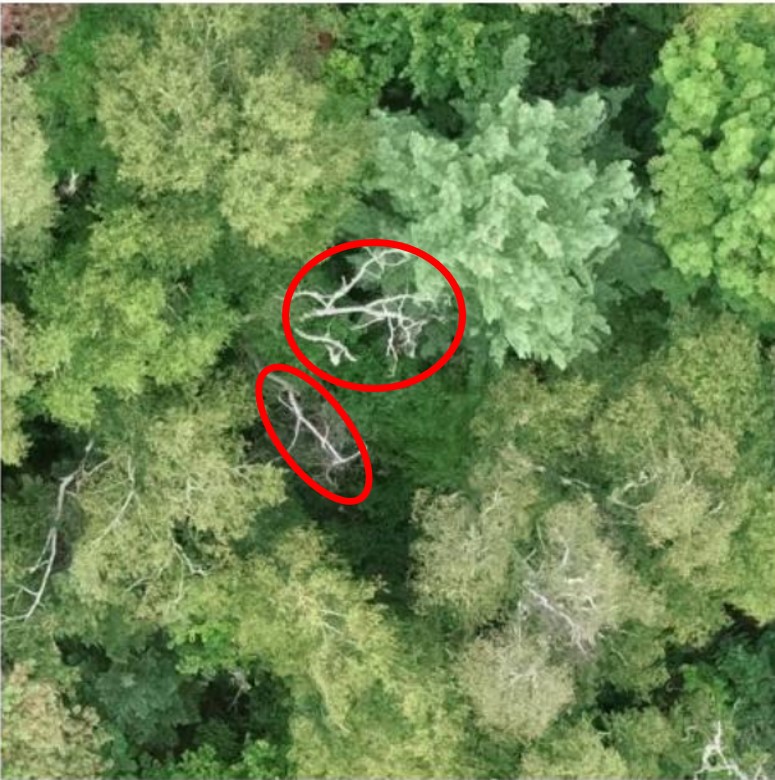}
        \end{subfigure} &
        \begin{subfigure}{\linewidth}
            \centering
            \includegraphics[width=.8\linewidth]{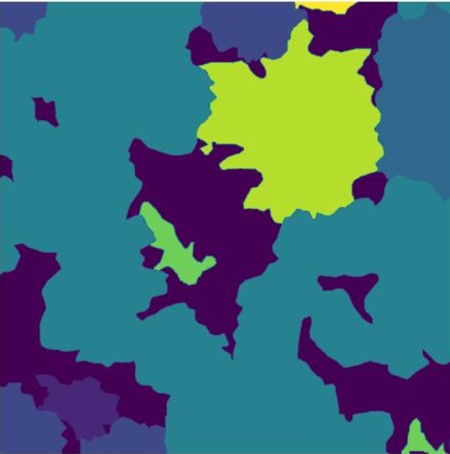}
        \end{subfigure} &
        \begin{subfigure}{\linewidth}
            \centering
            \includegraphics[width=.8\linewidth]{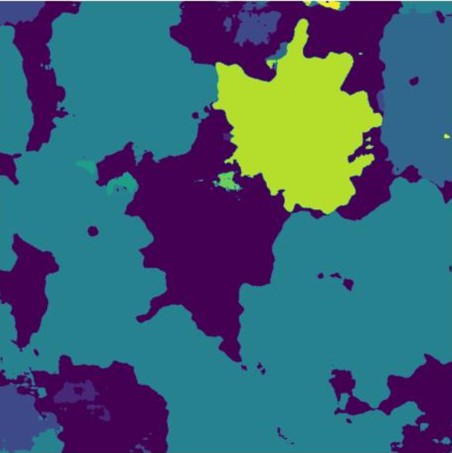}
        \end{subfigure} &
        \begin{subfigure}{\linewidth}
            \centering
            \includegraphics[width=.8\linewidth]{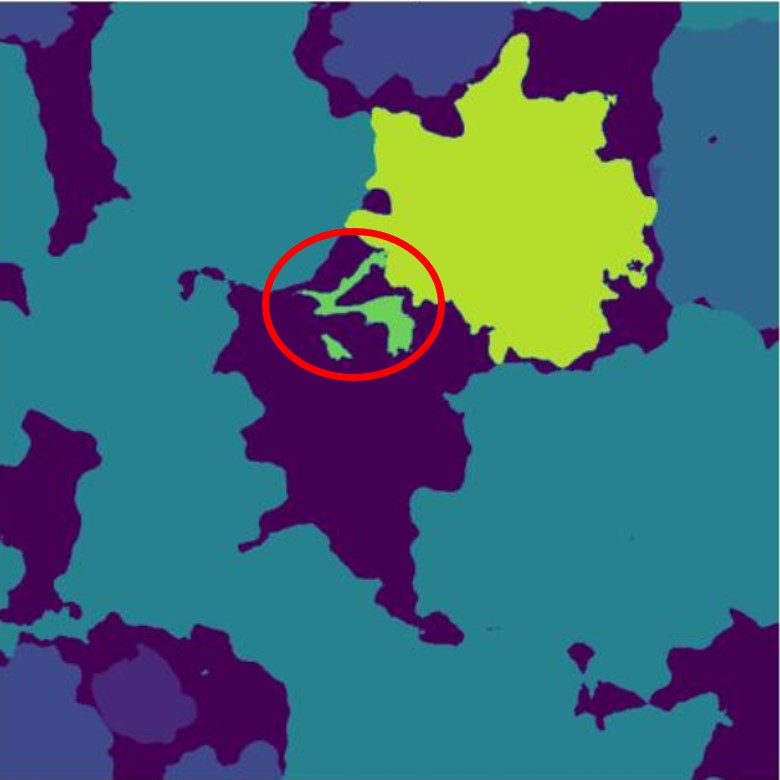}
        \end{subfigure} &
        \begin{subfigure}{\linewidth}
            \centering
            \includegraphics[width=.8\linewidth]{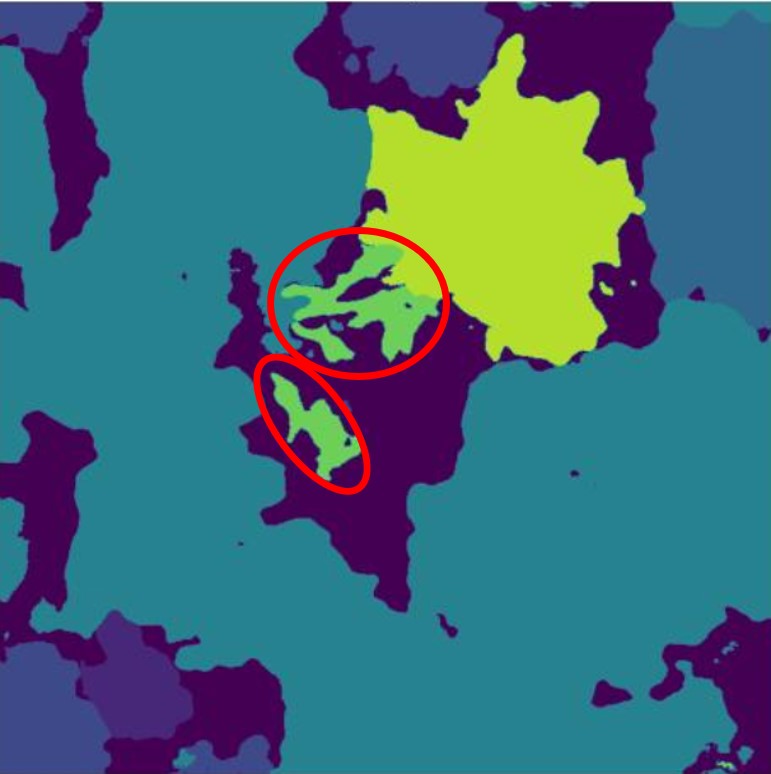}
        \end{subfigure} \\
        \bottomrule
    \end{tabularx}
    \caption{Aerial Tree Crown Semantic Segmentation of a Dense Forest (TreeCrown), Comprised of Visually Similar Tree Species. The left column shows the original image; the right columns show the labelled masks and the segmentation results. Red circles indicate dead trees.}
    \vspace{-10pt}
    \label{fig:QualitativeTreeCrown}
\end{figure*}

To assess the performance of our proposed segmentation model, we conduct a visual comparison of tree crown segmentation results across the original ResUNet and proposed PerceptiveNet (LogGab PerceptiveNet). Furthermore, to asses the effect of the proposed initial Log-Gabor parameterised convolutional layer, we provide a visual comparison of the proposed model with an initial parameterised Gabor convolutional layer (Gabor PerceptiveNet). The segmentation results, presented in Figure \ref{fig:QualitativeTreeCrown}, highlight each model's effectiveness in capturing tree crowns with varying characteristics, including living and decaying trees. While accuracy in segmenting tree crowns is the main focus, the red-circled dead trees aid analysis as reference points.


ResUNet demonstrates a reasonable ability to segment large tree crowns, yet it encounters challenges when it comes to finer boundary delineations, particularly with smaller and irregularly shaped trees. The model often over-segments larger crowns while under-segmenting smaller ones, leading to a reduction in segmentation accuracy. Dead trees, which exhibit sparse or absent foliage, are especially problematic for ResUNet, with frequent misclassifications or merging with nearby crowns observed in the results.

The Gabor PerceptiveNet, shows improved capability in capturing texture and boundary details. This is apparent in the clearer segmentation of tree crowns compared to ResUNet. While some inaccuracies persist—most notably with smaller crowns and more complex crown structures—the model provides improvements, particularly in separating adjacent tree crowns. Despite errors, the Gabor filters help the model differentiate between trees more effectively, including those exhibiting signs of decay or death.

PerceptiveNet exhibits the highest accuracy in segmenting tree crowns, particularly when it comes to delineating boundaries. The use of convolutional layer parameterised by trainable Log-Gabor functions enables the model to capture finer spatial details, translating into more precise segmentation across various tree crown sizes. PerceptiveNet excels at distinguishing closely situated trees and minimising over-segmentation, effectively handling complex crown shapes and smaller, irregular crowns. Notably, while dead trees pose a challenge for all models, PerceptiveNet outperforms them by capturing more high-frequency features, contributing to its superior segmentation performance.

Overall, PerceptiveNet delivers the most proficient segmentation of tree crowns, including those of varying sizes and health statuses. The Log-Gabor initial convolutional layer enhances feature extraction, allowing for clearer boundary detection and more accurate segmentation. These visualisations demonstrate the importance of the Log-Gabor properties: zero DC components, improved orthogonality across scales, uniform Fourier domain coverage, and superior spatial localisation in semantic segmentation.

\section{Conclusion}
\label{sec:conclusion}
While tree crown semantic segmentation is essential for ecology, forestry, agriculture, and biodiversity studies, it is significantly impacted by factors such as shadows, light variations, overlapping tree crowns, and weak distinctive features among tree species. These challenges are further exacerbated in dense forests during the green leaf season. To address these issues, we proposed PerceptiveNet, a model that extracts salient features while capturing contextual and spatial information through a wider receptive field. Our model significantly outperforms SOTA models on the TreeCrown and two benchmark aerial scene datasets. Qualitative analysis shows enhanced feature extraction, enabling clearer boundary detection and more accurate segmentation.

Additionally, quantitative and qualitative analyses demonstrate that the proposed convolutional layer, parameterised by trainable Log-Gabor functions, outperforms both traditional Gabor-based and standard convolutional layers by effectively leveraging the strengths of Log-Gabor filters for semantic segmentation. Moreover, an investigation into the proposed individual layers and their combinations reveals that while each contributes to performance gains, their synergistic integration leads to even greater improvements.

Finally, the proposed hybrid CNN-Transformer model, PerceptiveNeTr, illustrates the advantages of capturing long-range dependencies and global context. Although PerceptiveNeTr's performance metrics are lower compared to our pure CNN PerceptiveNet, it establishes a foundation for future research aimed at integrating advanced transformer models, such as the Hierarchical Vision Transformer, along with pre-trained ones to improve performance. Importantly, this comparison highlights the robust generalisation of our PerceptiveNet backbone across various applications.

\medskip

\noindent\textbf{Acknowledgements:} This work was funded in part by UKRI Frontiers grant EP/X024520/1.

{
    \small
    \bibliographystyle{ieeetr}
    \bibliography{main}
}


\end{document}